\documentclass[letterpaper, 10 pt, conference]{ieeeconf}  

\IEEEoverridecommandlockouts                              

\usepackage{amsmath}
\usepackage{booktabs}
\usepackage{multirow}
\usepackage{tabularx}
\usepackage{array}
\usepackage{graphicx}
\usepackage{amsfonts}
\usepackage{adjustbox}
\usepackage{xcolor}
\usepackage{cuted}
\usepackage{capt-of}

\usepackage[normalem]{ulem}

\title{\LARGE \bf
FPV-CRASH: A Dataset for Offline Crash Moment Localization \\ in Drone First-Person-View Videos
}

\author{
	Jifan Zhang$^{1,*}$,
	Tingyu Wang$^{1}$,
	Qian Liu$^{2}$,
	Zhedong Zheng$^{3}$%
	\thanks{$^{1}$School of Communication Engineering, Hangzhou Dianzi University, China.}%
	\thanks{$^{2}$School of Computer Science, Faculty of Science, University of Auckland, Auckland, New Zealand.}%
	\thanks{$^{3}$Department of Computer and Information Science, Faculty of Science and Technology, University of Macau, Macau, China.}%
	\thanks{$^{*}$Corresponding author.}%
}

\begin{document}

\bstctlcite{IEEEtranBSTControl}

\maketitle
\thispagestyle{empty}
\pagestyle{empty}

\begin{abstract}
Crash Moment Localization aims to localize the major physical impact that interrupts controlled flight in an onboard first-person-view (FPV) video recorded by a drone, thereby providing a temporal reference for retrospective crash analysis in aerial robotics.
However, existing crash-video datasets predominantly target surveillance or dash-camera footage, and the corresponding annotation schemes and task formulations are not designed for temporal localization of the major physical impact in UAV onboard FPV videos.
To address this gap, we present FPV-CRASH, a dataset and fixed-protocol reference evaluation for offline Crash Moment Localization in drone FPV videos. FPV-CRASH contains 1,052 drone FPV clips, with 526 Crash and 526 Normal clips; each Crash clip has a single-frame reference annotation for the Crash Moment, and every clip is paired with four automatically generated from sampled frames and the known Crash/Normal label English captions.
Normal clips carry no timestamp target and contribute training-time background supervision only. The captions support an auxiliary Text-to-Video Retrieval task, whereas localization inference remains video-only. We further provide a heatmap-based reference model that estimates a continuous Crash Moment timestamp from sparsely sampled frames.
Under the fixed clip-level protocol, the reference model reduces MAE from 1.12s for the strongest fixed timing-prior baseline to 0.56s (a 50\% relative reduction) and achieves a median absolute error (MedAE) of 0.33s; the fraction of clips localized within 0.5s rises from 37.14\% to 65.71\%. The release contains the annotations, available source video identifiers and URLs, temporal clipping information, fixed split lists, and preprocessing code so that the reportable portion can be reconstructed without redistributing the original source footage. At submission time, all recorded source-compilation URLs remained accessible.
\end{abstract}

\section{Introduction}

Unmanned aerial vehicles (UAVs) are increasingly used for visual data capture~\cite{zheng2023uavm}. Their first-person-view (FPV) videos record flight motion and environmental contact, but a Crash/Normal label does not identify when controlled flight is interrupted. Localizing the major physical impact provides a temporal anchor for retrospective FPV crash analysis.The resulting timestamp provides a boundary between pre-impact flight and post-impact evidence, supporting targeted review of variable-length crash clips.
Prior UAV research addresses high-speed flight, collision-driven navigation, online collision response, and accident forensics~\cite{delmerico2019uzhfpv,gandhi2017learning,wang2024air,editya2024visual}. Much of this work targets online operation; we instead locate the major impact in a fully observed FPV video. At inference, the model receives only a complete video and returns one continuous Crash Moment timestamp without a text query.

\begin{figure*}[t]
    \centering
    \includegraphics[width=\textwidth]{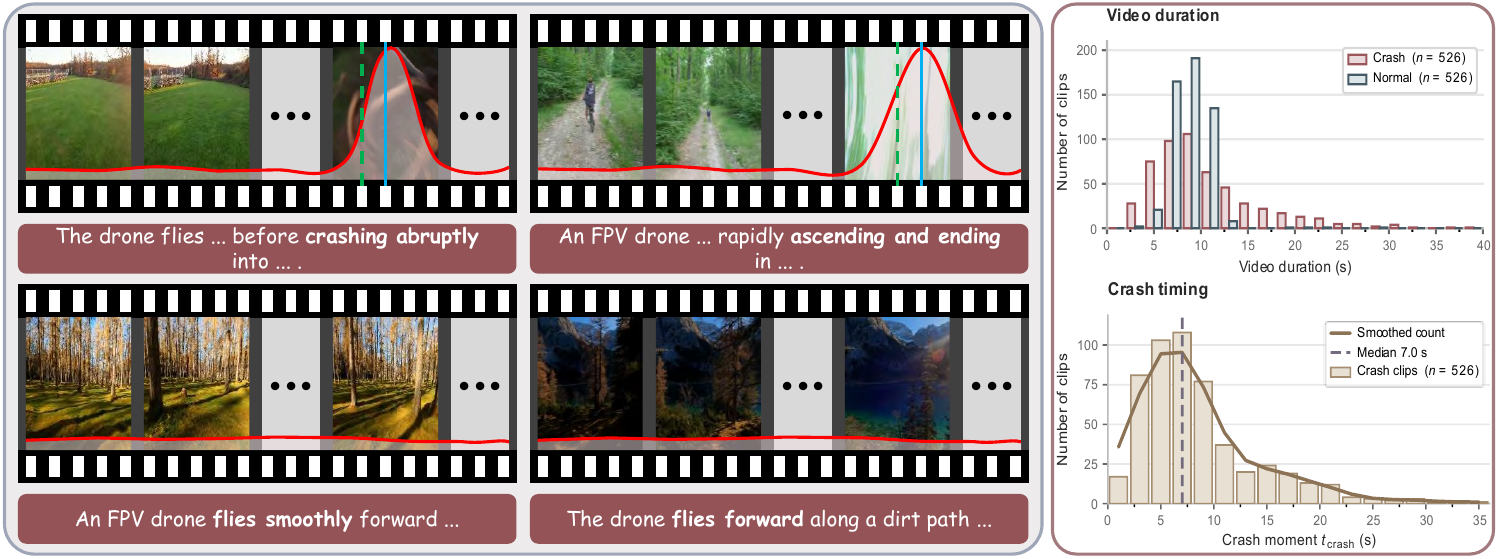}
    \vspace{-.2in}
    \caption{\textbf{Task definition and dataset statistics of FPV-CRASH.} Red curves are Crash responses; green dashed and blue solid lines denote reference and predicted moments. Normal clips have no Crash Moment; text strips show truncated captions. Right panels show Crash/Normal clip-duration and Crash Moment distributions in 2-s bins; the lower solid curve is smoothed and the dashed line marks the 7.0-s median.}
    \label{fig:crash_moment_localization}
    \vspace{-.1in}
\end{figure*}

FPV-CRASH contains 1,052 class-balanced clips, 4,208 captions, and single-frame Crash Moment references for 526 Crash clips. It uses a fixed class-stratified 60/20/20 split; localization (Fig.~\ref{fig:crash_moment_localization}) is evaluated on Crash clips, while Normal clips provide background supervision. Our reference model combines sparse sampling with temporal modeling to predict continuous timestamps; optional video--text alignment provides training-time supervision and auxiliary retrieval, while localization inference remains video-only. We evaluate positional encoding, training-time text supervision, and Normal background supervision. Our contributions are as follows:
\begin{figure}
    \centering
    \vspace{-.1in}
    \includegraphics[width=0.48\textwidth]{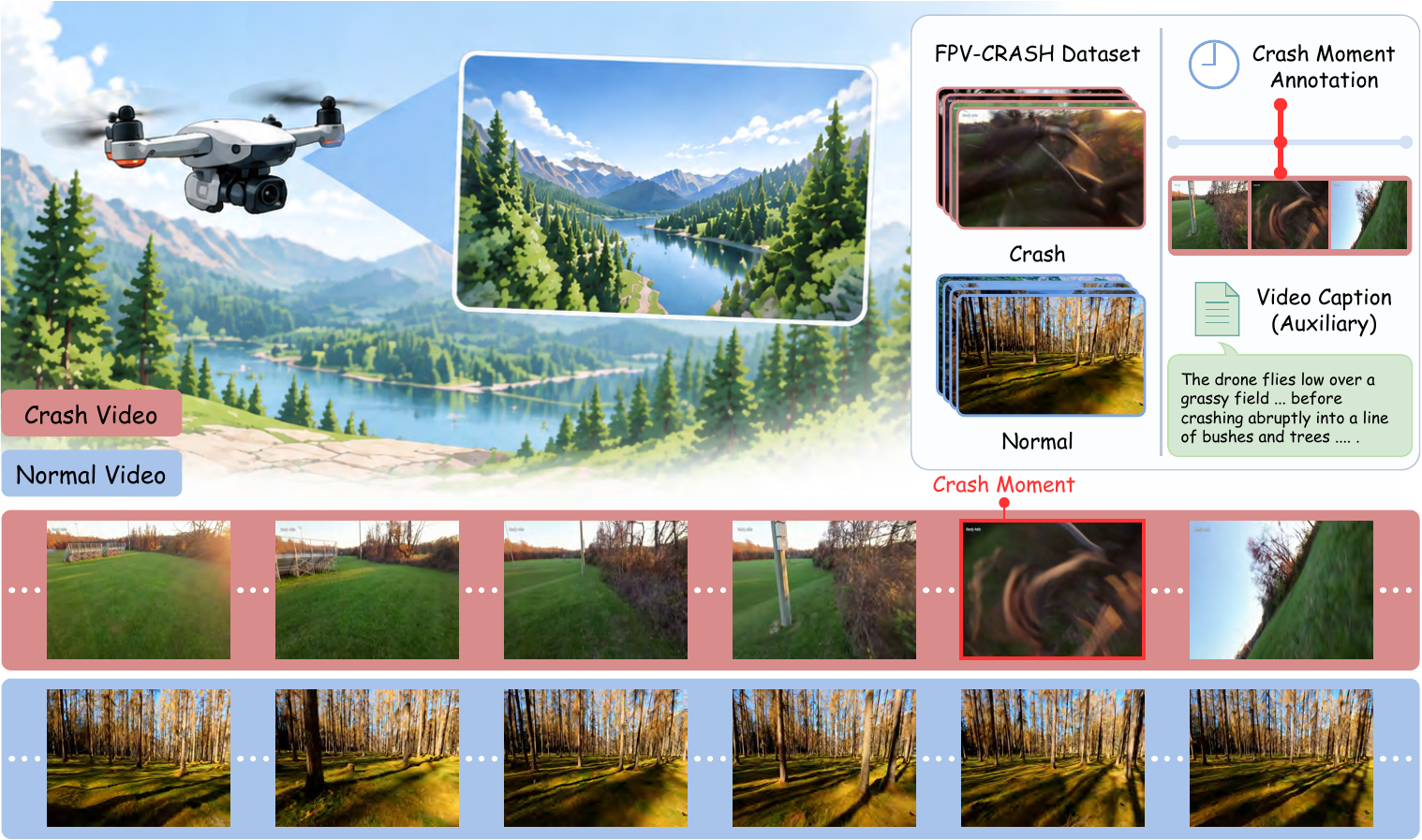}
    \vspace{-.1in}
    \captionof{figure}{
        \textbf{Overview of FPV-CRASH and Crash Moment Localization.}
        The dataset contains Crash and Normal FPV videos with Crash Moment annotations.
        Given a fully observed FPV video, the task is to localize the major physical
        impact that interrupts the flight.
    }
    \vspace{-.1in}
    \label{fig:task_overview}
\end{figure}

$\bullet$ We construct FPV-CRASH: 1,052 class-balanced FPV clips from approximately 40 crash compilations and three normal-flight compilations, with single-frame Crash Moment references and four automatically generated captions per clip (4,208 captions in total).

$\bullet$ We define Crash Moment Localization and its fixed evaluation protocol, and provide a heatmap-based reference model for continuous timestamp prediction.

$\bullet$ We examine positional encoding, training-time text supervision, and Normal background supervision under the fixed protocol, with detailed results discussed in Experiments. Among the controlled component ablations, the largest observed degradations occur when all heatmap BCE terms or slot encoding are removed, increasing MAE from 0.5585s to 0.8508s and 1.4221s, respectively.

\section{Related Work}

\noindent\textbf{UAV Visual Data and Autonomous Flight.}
Onboard UAV datasets cover state estimation and localization (EuRoC, Zurich Urban MAV~\cite{burri2016euroc,majdik2017zurich}), high-speed flight (UZH-FPV~\cite{delmerico2019uzhfpv}), and varied sensing conditions (INSANE~\cite{brommer2024insane}); Race Against the Machine covers autonomous and piloted racing~\cite{bosello2024ratm} and FWAF-VID aggressive flapping-wing flight~\cite{jiang2025fwafvid}, but none provides crash-centered frame-level annotations. UAV vision research instead studies drone--satellite matching~\cite{wang2022eachpart}, language-guided geo-localization~\cite{chu2024geotext}, tracking with semantic conditioning or hierarchical feature aggregation~\cite{blei2025cloudtrack,li2025cgtrack}, time-optimal trajectory planning~\cite{qin2024time}, and risk-aware UAV--UGV routing~\cite{mondal2025risk}---objectives that differ from retrospective localization of a physical crash impact.

\noindent\textbf{UAV Collision Avoidance, Detection, and Failure Understanding.}
Collision-oriented UAV research targets prevention or online response: Learning to Fly by Crashing learns navigation from collision experience~\cite{gandhi2017learning}, ColANet provides collision-avoidance data~\cite{pedro2020colanet}, and later methods use optical flow~\cite{pedro2021collision}, integrated perception, planning, and control~\cite{liu2025dynamic}, or vision-transformer policies~\cite{bhattacharya2025vision}. After-impact work includes Air Bumper's inertial collision detection, recovery, and mapping~\cite{wang2024air} and pre-/post-accident frame labels and explanations from FPV videos~\cite{editya2024visual}. Fault-detection methods flag anomalous states or component faults~\cite{keipour2019automatic,keipour2021alfa,le2025rflymad}, which need not coincide with physical impact. We instead localize the major impact interrupting controlled flight from a complete video, rather than predicting, avoiding, or diagnosing failures online.

\noindent\textbf{Temporal Event Localization.}
Temporal localization targets point or interval events: SoccerNet casts timestamp prediction as action spotting~\cite{giancola2018soccernet}, ActivityNet Captions associates events with described intervals~\cite{krishna2017dense}, and temporal action localization estimates action boundaries or intervals~\cite{lin2019bmn,zhang2022actionformer}. Natural-language moment retrieval conditions interval prediction on a query~\cite{zhang2020tan,lei2021qvhighlights}, with active learning reducing annotation cost~\cite{ji2023binary}. Ego4D introduces Moment Queries for activity-interval localization and a Point-of-No-Return task for state-change keyframe localization~\cite{grauman2022ego4d}. ACCIDENT localizes traffic impacts in fixed-view surveillance videos~\cite{picek2026accident}, while accident anticipation predicts events from partial observations~\cite{chan2016anticipating}. In contrast, our task returns one major physical-impact timestamp from a fully observed drone FPV video, without an inference-time query.

\noindent\textbf{Multimodal Representation Alignment.}
Contrastive vision--language learning aligns visual and textual embeddings~\cite{radford2021clip}, and Frozen in Time extends this to video~\cite{bain2021frozen}; CLIP4Clip studies temporal aggregation~\cite{luo2022clip4clip}, Align and Tell combines local alignment with captioning supervision~\cite{wang2023align}, and GRAM uses a Gramian-volume objective for multimodal alignment~\cite{cicchetti2025gramian}. Our implementation adapts the official GRAM codebase, retaining its encoder organization and contrastive projection heads but excluding GRAM's Gramian-volume and data--anchor objectives; we instead employ video--text alignment with a scalar temperature and add a Temporal Transformer, Gaussian heatmap target, temporal localization head, and masked-Softmax expectation decoder for Crash Moment prediction.

\section{FPV-CRASH Dataset}

FPV-CRASH provides continuous onboard FPV clips with Crash/Normal labels, single-frame Crash Moment annotations for Crash clips, and English captions.

\subsection{Crash Definition and Annotation Protocol}

A Crash requires visible or highly credible physical contact that causes a loss of controlled flight or terminates the flight. The Crash Moment is the major impact after which control is not recovered, not an earlier light contact. Crashes involving a briefly occluded impact, water contact, rebound, or short post-impact motion qualify when the collision and interruption remain highly credible. Uncontrolled descents, blackouts, frozen imagery, and abrupt endings without an observable or highly credible contact event are excluded. FPV-CRASH may therefore under-represent impacts with occluded, poorly illuminated, or missing contact evidence.

A primary annotator assigns each Crash clip a single-frame reference timestamp through repeated playback and frame-level inspection; ambiguous samples are excluded. The annotation records a frame index and its 30-fps timestamp. Because an impact may span adjacent frames, this is a temporal reference rather than a uniquely precise physical instant. A second annotator independently re-annotated 53 Crash clips (10\% of the Crash set); the two references fall within 0.5s on 85.7\% and within 1s on 95.2\% of these clips, and the two annotators agree on whether an annotatable impact is present with Cohen's $\kappa$ = 0.88.

\subsection{Data Collection and Curation}

Publicly accessible YouTube compilations provide the source material: approximately 40 crash compilations and three longer normal-flight compilations. The dataset is class-balanced but not source-balanced. We retain continuous single-drone FPV sequences and discard mixed viewpoints, adjacent edits, duplicates, and severe visual defects. Crash compilations routinely re-upload the same underlying footage under different edits, so per-compilation deduplication cannot catch reuse across sources: we therefore add an all-pairs near-duplicate scan over all 1,052 clips (CLIP-feature cosine similarity plus a perceptual hash) and report 13 pairs above the duplicate threshold. Because the split is clip-disjoint but not source-disjoint, a reused clip that lands in both Train and Test would let a model read the timestamp off a memorized instance, so the scan is a precondition for the headline MAE rather than a housekeeping step. We use content-change detection and manual refinement to segment Crash clips, retaining continuous pre-impact flight, the major collision, and sufficient post-impact evidence. Individually selected boundaries produce variable clip and pre-impact durations. For Normal clips, visual-change-based scene detection generates candidates; we sample away from edits, select durations broadly comparable to Crash clips, and screen for continuous flight without collisions or severe defects.

We standardize all 1,052 clips to 30 fps without audio while retaining their original spatial resolution, and manually review them for completeness and visual quality. Each clip has four English captions generated with \texttt{qwen3.7-plus} from Alibaba Cloud Model Studio using sampled frames and the known Crash/Normal class. Class-conditioned generation reduces gross class inconsistency but may make text-supervision effects partly reflect class separation. Captions contain no Crash Moment timestamp, provide only auxiliary training supervision, and are unnecessary for video-only localization inference. We check generation completion programmatically, but captions receive no formal human correctness audit and are not exhaustively reviewed or edited; therefore, they are excluded from semantic dataset statistics. All source footage is third-party, publicly accessible material. We do not redistribute video; we release identifiers, timestamps, and preprocessing code only, and we describe how removal requests are handled. The clips may depict identifiable persons, private property, or injury, which is why no per-clip source URL appears in the paper body. 

\subsection{Dataset Statistics and Splits}

Table~\ref{tab:dataset_stats} summarizes the 1,052 class-balanced clips, 4,208 captions, and fixed splits; only Crash clips have timestamp targets. Across 526 Crash clips, clip duration has a mean of 10.61s and a median of 9.23s (SD 5.83s; range 2.67--38.93s), while pre-impact duration has a mean of 8.47s and a median of 7.02s (SD 5.61s; range 0.87--35.23s). Normal clips average approximately 8.71s, so the classes are broadly, not strictly, duration-matched. For the 105 Crash test clips, the mean and median of $\tau_i/T_i$ are 0.767 and 0.796, and the mean post-impact duration is 2.16s. The fixed seed-50, clip-level, class-stratified 60/20/20 split contains no overlapping clips across partitions but is not source-disjoint because clips are not grouped by compilation or scene. We therefore also report a source-disjoint variant that holds out entire compilations, where the Crash MAE changes from 0.5585 to 0.7239s; this bounds how much of the reference result is source familiarity. Results therefore characterize performance under this protocol, not generalization to unseen sources, scenes, flights, or drone platforms. The supervision bandwidth is not constant across clips. With $\Delta_i \approx T_i/K$ at test time, $\sigma_i = \operatorname{clip}(0.5\Delta_i, 0.25, 2.5)$ grows linearly with clip duration, so $\sigma_i \approx 0.05\,T_i$ over the duration range observed here: 0.25s at the shortest clip, 0.46s at the median clip, 0.53s at the mean clip, and 1.95s at the longest. 

\begin{table}[t]
\centering
\caption{FPV-CRASH dataset composition and fixed splits.}
\label{tab:dataset_stats}
\footnotesize
\setlength{\tabcolsep}{5pt}
\renewcommand{\arraystretch}{1.08}
\begin{tabular}{l|cccc}
    \toprule
    Partition & Crash & Normal & Clips & Captions \\
    \midrule
    Train      & 316 & 316 & 632   & 2,528 \\
    Validation & 105 & 105 & 210   & 840   \\
    Test       & 105 & 105 & 210   & 840   \\
    \midrule
    \textbf{Total}
    & \textbf{526}
    & \textbf{526}
    & \textbf{1,052}
    & \textbf{4,208} \\
    \bottomrule
\end{tabular}
\vspace{-.1in}
\end{table}

\subsection{Comparison with Existing Datasets}

Table~\ref{tab:dataset_comparison} compares representative UAV datasets
and temporal-video benchmarks. Scale uses source-specific units and is
not directly comparable across rows: Learning to Fly by Crashing counts
collision trajectories, FWAF-VID reports its 17 flight sequences
excluding auxiliary static-flapping and calibration recordings, and the rounded total for ACCIDENT includes both real CCTV and synthetic clips.

In the Modality column, ``FPV'' denotes an explicitly identified
first-person view, whereas ``onboard'' more generally denotes
platform-mounted camera data. ``ColANet-derived'' refers to the
100 ColANet videos processed and annotated by Editya \emph{et al}. 	The ``Normal'' column denotes independently curated Normal clips, not
background portions within event videos; ``--'' denotes an absent or
inapplicable entry. FPV-CRASH combines balanced Crash/Normal FPV clips,
single-frame Crash Moment references, and automatically generated captions.

\begin{table*}[t]
\centering
\caption{Comparison with representative onboard UAV and temporal video datasets.}
\label{tab:dataset_comparison}
\vspace{-.1in}
\footnotesize
\setlength{\tabcolsep}{2.2pt}
\renewcommand{\arraystretch}{0.98}

\begin{adjustbox}{max width=\textwidth}
    \begin{tabular}{@{}lclllcl@{}}
        \toprule
        Dataset & Scale & Modality & Task & Time ann. & Normal & Text ann. \\
        \midrule
        
        Learning to Fly by Crashing~\cite{gandhi2017learning} &
        11.5K traj. &
        Onboard video + IMU &
        Collision avoidance &
        IMU-derived collision time &
        No &
        -- \\
        
        UZH-FPV~\cite{delmerico2019uzhfpv} &
        27+ seq. &
        FPV video + IMU + events &
        State estimation &
        -- &
        -- &
        -- \\
        
        ColANet-derived~\cite{editya2024visual} &
        100 videos &
        FPV video &
        Accident forensics &
        Accident time + frame labels &
        No &
        Frame explanations \\
        
        ACCIDENT~\cite{picek2026accident} &
        4.2K clips &
        Traffic CCTV &
        Accident localization &
        Impact time &
        No &
        -- \\
        
        Race Against the Machine~\cite{bosello2024ratm} &
        30 flights &
        Onboard video + IMU &
        Drone racing &
        -- &
        -- &
        -- \\
        
        FWAF-VID~\cite{jiang2025fwafvid} &
        17 seq. &
        Onboard cameras + IMU &
        Visual-inertial localization &
        -- &
        -- &
        -- \\
        
        SoccerNet~\cite{giancola2018soccernet} &
        500 games &
        Broadcast video &
        Action spotting &
        1-s event time &
        No &
        -- \\
        
        ActivityNet Captions~\cite{krishna2017dense} &
        20K videos &
        Web video &
        Dense captioning &
        Start--end intervals &
        No &
        Event descriptions \\
        
        \addlinespace[1pt]
        
        \textbf{FPV-CRASH (ours)} &
        \textbf{1,052 clips} &
        \textbf{FPV video} &
        \textbf{Crash Moment Localization} &
        \textbf{Frame-level Crash Moment} &
        \textbf{Yes (526)} &
        \textbf{4 auto captions/clip} \\
        
        \bottomrule
    \end{tabular}
\end{adjustbox}
\vspace{-.1in}
\end{table*}

\section{Method}


\begin{figure*}[t]
    \centering
    \includegraphics[width=\textwidth]
{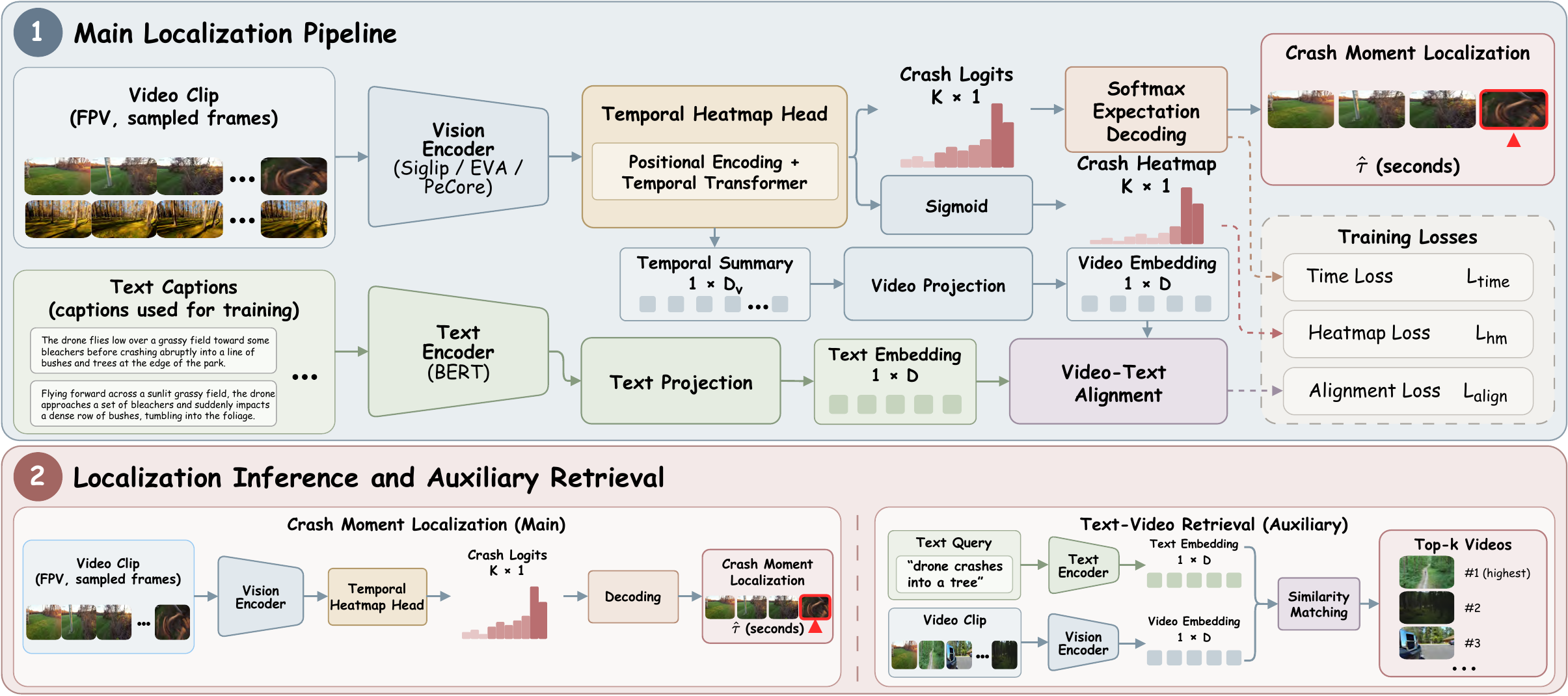}
    \vspace{-.2in}
    \caption{\textbf{The proposed heatmap-based localization framework with auxiliary video--text alignment.} Temporal video features predict a Crash heatmap and continuous timestamp; text-video alignment is used only during training, while localization inference is video-only. The shared embedding space also supports Text-to-Video Retrieval.}
    \label{fig:framework}
    \vspace{-.1in}
\end{figure*}

\textbf{Task Definition.} For video $V_i$ of duration $T_i$, $y_i\in\{0,1\}$ denotes Normal or Crash. A Crash video has a reference Crash Moment timestamp $\tau_i$; a Normal video has no timestamp target. Given a complete clip, the model returns one continuous timestamp, evaluated only on Crash videos. The frame-index sequence is divided into $K$ contiguous segments with approximately equal frame counts. We sample one frame uniformly from each segment during training and select its center frame during validation and testing. Sampling times are computed from the selected frame indices and the video frame rate. For slot $k$, $t_{ik}$, $m_{ik}\in\{0,1\}$, and $\mathbf f_{ik}$ denote sampling time, validity, and sampled-frame feature. Invalid slots are masked throughout. Positional encoding represents slot order; $t_{ik}$ constructs targets and decodes timestamps.

\subsection{Model Architecture and Temporal Encoding}

Building on the encoder architecture and contrastive projection heads of GRAM~\cite{cicchetti2025gramian}, we add a task-specific temporal branch for Crash heatmap prediction and masked-Softmax expectation decoding, without using the Gramian-volume or data--anchor objectives of GRAM. Frozen backbone features form a $K\times D_{\mathrm{enc}}$ sequence and are projected to a common temporal space. The Temporal Heatmap Head adds slot-indexed sinusoidal positional encoding~\cite{vaswani2017attention} and applies a non-causal Temporal Transformer:
\begin{equation}
    \mathbf u_{ik} = \phi(\mathbf f_{ik})+\mathbf e_k, 
    \mathbf H_i = \mathcal T([\mathbf u_{i1},\ldots,\mathbf u_{iK}],\mathbf m_i),
    \label{eq:temporal_encoding}
\end{equation}
where $\phi$ is the input projection, $\mathbf e_k$ the positional encoding, and $\mathbf H_i=[\mathbf h_{i1},\ldots,\mathbf h_{iK}]^\top$. A position-wise linear layer yields Crash logits $a_{ik}=\mathbf w^\top\mathbf h_{ik}+b$ for both heatmap supervision and timestamp decoding; there is no separate time-regression head. For alignment, we use a layer-normalized masked mean of the valid outputs as the temporal summary. The temporal summary and the \texttt{[CLS]} output of a trainable BERT~\cite{devlin2019bert} encoder are then projected and normalized in a shared space:
\begin{equation}
    \begin{aligned}
        \mathbf z_i^v &= \operatorname{norm}(g_v(\bar{\mathbf h}_i)), \bar{\mathbf h}_i = \operatorname{LN}\!\left(\frac{\sum_k m_{ik}\mathbf h_{ik}}{\sum_k m_{ik}}\right),\\
        \mathbf z_i^t &= \operatorname{norm}\big(g_t(f_t(C_i)_{\mathrm{[CLS]}})\big).
    \end{aligned}
    \label{eq:cross_modal_embedding}
\end{equation}

The temporal summary supports alignment and retrieval but does not directly enter Crash Moment decoding.

\subsection{Crash Heatmap and Timestamp Decoding}

A Gaussian target maps each Crash reference to sampled slots. With $\Delta_i$ the median positive gap between adjacent valid sampling times, Crash targets are defined as:
\begin{equation}
    \begin{gathered}
        \sigma_i
        = \operatorname{clip}(\alpha\Delta_i,\,\sigma_{\min},\,\sigma_{\max}), \\[2pt]
        \tilde q_{ik}
        = m_{ik}\exp\!\left[-\frac{(t_{ik}-\tau_i)^2}{2\sigma_i^2}\right],
        \qquad
        q_{ik}
        = \frac{\tilde q_{ik}}{\max_j \tilde q_{ij}}.
    \end{gathered}
    \label{eq:heatmap_target}
\end{equation}

Valid Normal positions receive zero. The sigmoid response $s_{ik}=\operatorname{sigmoid}(a_{ik})$ forms the supervised and visualized Crash heatmap. Separately, masked softmax normalizes the same logits only over valid slots, and expectation decoding returns the continuous timestamp:
\begin{equation}
    p_{ik} = \frac{m_{ik}\exp(a_{ik})}{\sum_j m_{ij}\exp(a_{ij})}, 
    \hat\tau_i = \sum_k p_{ik}t_{ik}.
    \label{eq:timestamp_decoding}
\end{equation}

Thus $\mathbf S_i=[s_{i1},\ldots,s_{iK}]^\top$ contains position-wise, uncalibrated Crash responses, whereas $\mathbf P_i$ is a temporal decoding distribution over valid sampled slots. The expectation in Equation~\eqref{eq:timestamp_decoding} yields a continuous timestamp~\cite{sun2018integral}. Both outputs derive from the same Crash logits, but only the temporal decoding distribution is normalized across sampling positions. Heatmap responses provide supervision and visualization; the decoding distribution determines the timestamp. A Normal-video output is not a Crash Moment and is excluded from localization metrics.

\subsection{Training Objective}

Let $\mathcal C$ and $\mathcal N$ be the Crash and Normal subsets of a batch and $M_i=\sum_k m_{ik}$. Binary cross-entropy defines
\begin{equation}
    \begin{aligned}
        \mathcal L_{\mathrm{heat}}^C
        &=\frac{1}{|\mathcal C|}\sum_{i\in\mathcal C}\frac{1}{M_i}
        \sum_k m_{ik}\ell_{\mathrm{bce}}(a_{ik},q_{ik}),\\
        \mathcal L_{\mathrm{heat}}^N
        &=\frac{1}{|\mathcal N|}\sum_{i\in\mathcal N}\frac{1}{M_i}
        \sum_k m_{ik}\ell_{\mathrm{bce}}(a_{ik},0),\\
        \mathcal L_{\mathrm{hm}}&=w_h\left(\mathcal L_{\mathrm{heat}}^C+w_n\mathcal L_{\mathrm{heat}}^N\right),
    \end{aligned}
    \label{eq:heatmap_loss}
\end{equation}
where $w_h$ weights heatmap supervision and $w_n$ controls the relative contribution of Normal background supervision. Normal clips have no timestamp target. For Crash videos, the timestamp loss $\mathcal L_{\mathrm{time}}$ is the Smooth L1 loss between normalized predicted and reference timestamps:
\begin{equation}
    \mathcal L_{\mathrm{time}}=
    \frac{1}{|\mathcal C|}\sum_{i\in\mathcal C}
    \ell_{\mathrm{sL1}}\!\left(\frac{\hat\tau_i}{T_i},\frac{\tau_i}{T_i}\right).
    \label{eq:time_loss}
\end{equation}

With captions, matched video--text pairs serve as positives in a symmetric contrastive objective~\cite{radford2021clip}. For normalized embeddings, $r_{ij}=(\mathbf z_i^t)^\top\mathbf z_j^v/\theta$, where $\theta$ is a learned scalar temperature.	
\begin{equation}
    \mathcal L_{\mathrm{t2v}} = \frac{1}{B}\sum_i \ell_{\mathrm{ce}}^{\epsilon}(\mathbf r_{i:},i), 
    \mathcal L_{\mathrm{v2t}} = \frac{1}{B}\sum_i \ell_{\mathrm{ce}}^{\epsilon}(\mathbf r_{:i},i).	
    \label{eq:alignment_loss}
\end{equation}

Here $\ell_{\mathrm{ce}}^{\epsilon}$ is label-smoothed cross-entropy and $\mathcal L_{\mathrm{align}}=\tfrac12(\mathcal L_{\mathrm{t2v}}+\mathcal L_{\mathrm{v2t}})$. The complete objective is
\begin{equation}
    \mathcal L=\mathcal L_{\mathrm{hm}}+w_t\mathcal L_{\mathrm{time}}+w_a\mathcal L_{\mathrm{align}}.
    \label{eq:overall_loss}
\end{equation}

The aligned $L_2$-normalized embeddings support bidirectional retrieval. Given a normalized text query, videos in a precomputed candidate index are ranked by cosine similarity; video-to-text retrieval reverses the roles.

\section{Experiments}


\noindent\textbf{Implementation details.}
Unless otherwise specified, we use frozen SigLIP2 ViT-B/16~\cite{tschannen2025siglip2} and cache all frame features; backbone comparisons additionally evaluate frozen EVA01-CLIP-g/14~\cite{fang2023eva} and PE-Core ViT-B/16~\cite{bolya2025perception}, holding the split, temporal architecture, sampling protocol, and prediction head fixed. Frames are resized to $224\times224$ without spatial or photometric augmentation: SigLIP2 and PE-Core use bicubic interpolation and $[-1,1]$ normalization, EVA01 bilinear interpolation and CLIP normalization. We use $K=10$ segments, random within-segment samples for training and segment-center samples for validation and testing; frame rounding limits the within-video difference between adjacent test-slot spacings to about 0.033s, and substituting uniform random slot sampling for segment centres at test time changes MAE from 0.5585s to 0.6239s. The two-layer Temporal Transformer has dimension 512, eight heads, and dropout 0.1, with 512-dimensional video and text projections; BERT-base-uncased~\cite{devlin2019bert} is capped at 40 tokens, and each training video uses one of four randomly sampled captions. Gaussian parameters are $\alpha=0.5$, $\sigma_{\min}=0.25$s, and $\sigma_{\max}=2.5$s; Crash heatmap, Normal background, timestamp, and alignment weights are 1, 0.25, 1, and 0.5. Since $\sigma$ sets the blur scale of the only dense supervision signal, we sweep $\alpha\in\{0.25,0.5,1.0\}$ and additionally evaluate a duration-normalized variant predicting in $\tau/T$ with constant normalized $\sigma$, which removes the coupling between supervision width and clip length. The learned temperature starts at 0.07, is lower-bounded by $10^{-4}$, and label smoothing is $\epsilon=0.1$. Single-GPU main runs use 20 epochs, global batch size 16, and drop-last batching; a Crash-only control extends to 40 epochs, otherwise unchanged. The step count matters: 39 updates per epoch on 632 clips give 780 total updates for the full model, small for a trainable BERT encoder plus the temporal branch. We therefore report training and validation curves and confirm validation Crash MAE has plateaued before recording any table entry, since comparisons taken before convergence measure the optimiser trajectory rather than the architecture. AdamW uses $(\beta_1,\beta_2)=(0.9,0.98)$, weight decay 0.01, learning rates $1.25\times10^{-4}$ for the main model and $10^{-4}$ for BERT, and 10\% linear warmup followed by cosine decay.

\noindent\textbf{Protocol and metrics.}
We evaluate all 105 Crash test videos from the fixed split without classification filtering; Normal videos have no localization target. We report mean absolute error (MAE), median absolute error (MedAE), and Within@$\delta$ (W@$\delta$), the percentage localized within $\delta$ seconds. P90 is the absolute-error threshold covering approximately 90\% of Crash test clips; lower is better. Each configuration is trained with three seeds (50, 51, 52) and we report the mean with a bootstrap 95\% confidence interval over the 105 test clips. At this sample size the approximate standard errors are 0.06--0.09s for MAE and 4.6 percentage points for W@0.5, so we mark MAE differences below 0.12s and W@0.5 differences below 9 points as unresolved rather than as gains. No claim of external generalization is made. For both localization and retrieval, we select checkpoints based on minimum validation Crash MAE; retrieval metrics do not select checkpoints. Auxiliary bidirectional retrieval is evaluated on 210 test videos, using the first caption in caption-index order for each video. Each direction contains 210 queries and 210 candidates, and only the exact video--caption pair is positive. We report Recall@K for this instance-level paired-caption retrieval task.

\noindent\textbf{Model configurations.}
PE denotes sinusoidal slot encoding; PE-only omits caption supervision. For Crash-and-Normal training, Full denotes the PE+Text configuration with the complete objective; the SigLIP2 Full model is the primary reference. No-PE variants retain the Temporal Transformer. Crash-and-Normal uses both classes; Crash-only uses Crash videos. We compare backbones with Crash-and-Normal training, then use SigLIP2 for controlled ablations. SigLIP2 Crash-and-Normal without PE or text is the learned baseline; its Full model serves as the reference for loss ablations. Training without text sets $\mathcal L_{\mathrm{align}}=0$; Crash-only training omits $\mathcal L_{\mathrm{heat}}^N$. In loss ablations, $w_h=0$ removes both heatmap terms, and $w_n=0$ removes only Normal background supervision.

\subsection{Reference Localization Results}

We evaluate three zero-training timing priors on the same 105 Crash test clips. These priors provide a reference for localization based on clip duration and the temporal annotation distribution, without visual evidence of the impact. For Crash samples in the training set, let $r_i=\tau_i/T_i$. The train-relative-median and train-relative-mean priors predict $T\operatorname{median}_{\mathrm{train}}(r_i)$ and $T\operatorname{mean}_{\mathrm{train}}(r_i)$, respectively; the corresponding training-set ratios are 0.795941 and 0.760723. The video-center prior predicts $0.5T$. None of these priors uses frames, captions, or model predictions. The train-relative-median prior has the lowest MAE among the three priors in Table~\ref{tab:timing_prior_baselines} and serves as the reference for the sample-level comparison. Relative to this prior, Full reduces MAE from 1.1159s to 0.5585s and MedAE from 0.6610s to 0.3310s. W@0.5 rises from 37.14\% to 65.71\%, W@1 rises from 58.10\% to 85.71\%, and P90 falls from 2.88s to 1.17s. Under the fixed split, these differences suggest that Full is not limited to reproducing the dominant relative-position prior.

\begin{table}[t]
    \caption{Zero-training timing-prior baselines on the 105 Crash
        test clips. Errors are in seconds.}
    \label{tab:timing_prior_baselines}
    \centering
    \footnotesize
    \setlength{\tabcolsep}{2.7pt}
    \begin{adjustbox}{max width=\columnwidth}
        \begin{tabular}{lrrrrrr}
            \toprule
            Baseline & MAE$\downarrow$ & MedAE$\downarrow$ & P90$\downarrow$
            & W@0.5$\uparrow$ & W@1$\uparrow$ & W@2$\uparrow$ \\
            \midrule
            Train-relative median
            & 1.1159 & 0.6610 & 2.8769 & 37.14\% & 58.10\% & 82.86\% \\
            Train-relative mean
            & 1.2666 & 0.8347 & 3.5206 & 38.10\% & 59.05\% & 80.95\% \\
            Video center
            & 3.4192 & 2.5500 & 7.7333 & 10.48\% & 17.14\% & 38.10\% \\
            \bottomrule
        \end{tabular}
    \end{adjustbox}
    \vspace{-.1in}
\end{table}

\begin{figure}[t]
    \centering
    \includegraphics[width=\columnwidth]
    {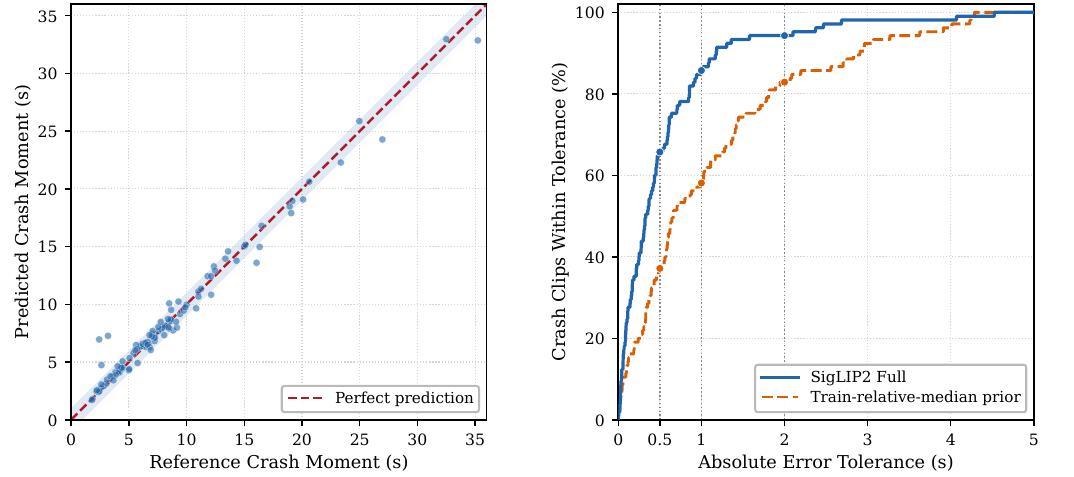}
    \vspace{-.25in}
    \caption{\textbf{Localization results on the 105 Crash clips in the fixed
        test split.} Top: SigLIP2 Full predictions versus reference Crash
        Moment timestamps; the dashed line denotes exact prediction and
        the shaded band indicates a $\pm 1$-s tolerance. Bottom: empirical
        absolute-error CDFs for SigLIP2 Full and the train-relative-median
        prior; vertical dotted lines mark 0.5-, 1-, and 2-s tolerances.}
    \label{fig:localization_results}
    \vspace{-.1in}
\end{figure}

Fig.~\ref{fig:localization_results} shows Full predictions against reference timestamps and the empirical absolute-error distributions. Full predictions broadly follow the identity line across the observed timestamp range, while their error distribution is shifted toward smaller errors than that of the train-relative-median prior. Several larger errors nevertheless remain in the distribution tail, consistent with Full having a 0.5585-s MAE that exceeds its 0.3310-s MedAE. Absolute error has descriptive Spearman correlations with clip duration ($\rho=0.348$, $\rho^2=0.12$) and pre-impact duration ($\rho=0.279$, $\rho^2=0.08$). The squared coefficients indicate modest rank associations rather than regression variance explained by duration alone, and the duration effect cannot be separated from the target construction in this analysis: because $\sigma_i \approx 0.05\,T_i$, longer clips receive proportionally wider supervision kernels under the current formulation. A duration-normalized formulation that predicts in $\tau/T$ with a constant normalized $\sigma$ would decouple these factors and provide a more direct test of their respective effects.

\subsection{Backbone Comparison}

Table~\ref{tab:backbone_localization} compares frozen backbones with Crash-and-Normal training and PE, with and without captions. EVA01-CLIP-g/14 is roughly an order of magnitude larger than the two ViT-B/16 backbones, so the 0.0172-s MAE spread across these rows mixes backbone scale with backbone identity. Full-model MAEs span 0.5516--0.5688s (0.0172s), and EVA01 yields the lowest MAE in the observed runs. SigLIP2 has the highest W@0.5, whereas PE-Core has the lowest MedAE among the Full models. No backbone leads on all reported localization metrics in these runs. The small observed spread does not establish a reliable backbone difference.

\begin{table}[t]
    \caption{Backbone comparison under Crash-and-Normal training. All models use PE; Text denotes training-time text supervision through video--text alignment. Errors are in seconds.}
    \label{tab:backbone_localization}
    \centering
    \footnotesize
    \setlength{\tabcolsep}{3pt}
    \begin{tabular}{llrrrr}
        \toprule
        Backbone & Text & MAE$\downarrow$ & MedAE$\downarrow$ &
        W@0.5$\uparrow$ & W@1$\uparrow$ \\
        \midrule
        EVA01
        & $\times$ & 0.5785 & 0.3753 & 58.10\% & 85.71\% \\
        & $\checkmark$ & \textbf{0.5516} & 0.3290 & 63.81\% & \textbf{87.62\%} \\
        \midrule
        SigLIP2
        & $\times$ & 0.6012 & 0.3808 & 60.00\% & 83.81\% \\
        & $\checkmark$ & 0.5585 & 0.3310 & \textbf{65.71\%} & 85.71\% \\
        \midrule
        PE-Core
        & $\times$ & 0.6227 & 0.3990 & 59.05\% & 85.71\% \\
        & $\checkmark$ & 0.5688 & \textbf{0.3214} & 64.76\% & \textbf{87.62\%} \\
        \bottomrule
    \end{tabular}
    \vspace{-.1in}
\end{table}

\subsection{Ablation Studies}

\noindent\textbf{Effect of primary components.}
Table~\ref{tab:core_localization} summarizes results for all PE/text combinations with Crash-and-Normal and Crash-only training. Removing PE retains the non-causal Temporal Transformer but removes explicit slot order, so it is a structural sanity check. PE lowers MAE from 1.4221/1.2792s (w/o and w/ text) to 0.6012/0.5585s for Crash-and-Normal, and from 1.2009/1.3446s to 0.6292/0.6494s for Crash-only. 

With PE, MAE changes from 0.6012s to 0.5585s for Crash-and-Normal but from 0.6292s to 0.6494s for 20-epoch Crash-only---opposite-direction results across training compositions that do not establish a stable caption/timing gain. Because alignment affects timestamp prediction only indirectly through gradients on the shared temporal trunk, these results, together with the step-matched control in Table~\ref{tab:step_matched_control}, are better interpreted as an auxiliary regularization effect than as evidence of dedicated localization supervision.

\begin{table}[t]
    \caption{SigLIP2 localization ablations. All variants retain the Temporal Transformer. Bold indicates the best result within each training block. }
    \label{tab:core_localization}
    \centering
    \footnotesize
    \setlength{\tabcolsep}{2.5pt}
    \begin{tabular}{lllrrrr}
        \toprule
        Training & PE & Text & MAE$\downarrow$ & MedAE$\downarrow$ &
        W@0.5$\uparrow$ & W@1$\uparrow$ \\
        \midrule
        Crash+Normal
        & $\times$ & $\times$ & 1.4221 & 0.6886 & 36.19\% & 65.71\% \\
        & $\checkmark$ & $\times$ & 0.6012 & 0.3808 & 60.00\% & 83.81\% \\
        & $\checkmark$ & $\checkmark$ & \textbf{0.5585} & \textbf{0.3310} & \textbf{65.71\%} & \textbf{85.71\%} \\
        & $\times$ & $\checkmark$ & 1.2792 & 0.6298 & 40.00\% & 67.62\% \\
        \midrule
        Crash-only
        & $\times$ & $\times$ & 1.2009 & 0.5712 & 43.81\% & 67.62\% \\
        & $\checkmark$ & $\times$ & \textbf{0.6292} & \textbf{0.3438} & \textbf{60.00\%} & 80.00\% \\
        & $\checkmark$ & $\checkmark$ & 0.6494 & 0.3628 & \textbf{60.00\%} & \textbf{80.95\%} \\
        & $\times$ & $\checkmark$ & 1.3446 & 0.6800 & 40.95\% & 61.90\% \\
        \bottomrule
    \end{tabular}
    \vspace{-.1in}
\end{table}

Crash-and-Normal and Crash-only training differ in optimizer-step count as well as data composition. The 20-epoch runs contain 780 and 380 updates, respectively. To examine this training-budget difference, the 40-epoch Crash-only control in Table~\ref{tab:step_matched_control} contains 760 updates, providing an approximately optimizer-step-matched comparison.

\begin{table}[t]
    \caption{SigLIP2 Crash-only 40-epoch control (760 updates), approximately step-matched to 20-epoch Crash-and-Normal training (780 updates). Errors are in seconds.}
    \label{tab:step_matched_control}
    \centering
    \footnotesize
    \setlength{\tabcolsep}{2.5pt}
    \begin{tabular}{llrrrr}
        \toprule
        PE & Text & MAE$\downarrow$ & MedAE$\downarrow$ & W@0.5$\uparrow$ & W@1$\uparrow$ \\
        \midrule
        $\times$ & $\times$ & 1.1540 & 0.6272 & 40.95\% & 69.52\% \\
        $\checkmark$ & $\times$ & \textbf{0.5879} & 0.3753 & \textbf{59.05\%} & \textbf{86.67\%} \\
        $\checkmark$ & $\checkmark$ & 0.7089 & \textbf{0.3418} & \textbf{59.05\%} & 80.00\% \\
        $\times$ & $\checkmark$ & 1.3440 & 0.5846 & 38.10\% & 64.76\% \\
        \bottomrule
    \end{tabular}
    \vspace{-.1in}
\end{table}

The extended Crash-only runs in Table~\ref{tab:step_matched_control} change results modestly and non-uniformly. The approximately step-matched PE-only MAE is 0.5879s, versus 0.6012s with Crash-and-Normal, so Normal clips do not uniformly improve localization. With text, Crash-only training yields 0.7089s versus 0.5585s for Crash-and-Normal. Because Normal clips alter both background supervision and alignment candidates, this comparison cannot isolate the Normal-background term.

\begingroup
\widowpenalty=10000
\noindent\textbf{Training objectives.}
Table~\ref{tab:loss_ablation} summarizes objective ablation results for SigLIP2 Full with Crash-and-Normal training. Removing Normal-background BCE retains Normal videos and their contribution to video--text alignment, so this comparison isolates the background loss term. MAE changes from 0.5585s to 0.5900s and W@0.5 from 65.71\% to 57.14\%. Normal mean/peak responses also rise from 0.000452/0.001815 to 0.160704/0.652930. These changes indicate that the term suppresses background responses and improves fine-tolerance localization in this run. With Normal-background BCE disabled, additionally removing Crash heatmap BCE raises MAE to 0.8508s. This isolates its conditional effect; removing all heatmap BCE is not a single-component ablation relative to the complete objective. Removing timestamp loss has a limited, metric-dependent effect (MAE 0.5585$\to$0.5607s; MedAE 0.3310$\to$0.3157s), while Softmax expectation decoding remains unchanged. These comparisons are descriptive under the fixed protocol.
\par
\endgroup

\begin{table}[t]
    \caption{Loss ablations for SigLIP2 Full with Crash-and-Normal training. The w/o all heatmap BCE variant removes both Crash and Normal heatmap terms; other variants remove the named component, and w/o alignment uses the PE-only configuration from Table~\ref{tab:core_localization}.}
    \label{tab:loss_ablation}
    \centering
    \footnotesize
    \setlength{\tabcolsep}{3pt}
    \begin{tabular}{lrrrr}
        \toprule
        Variant & MAE$\downarrow$ & MedAE$\downarrow$ &
        W@0.5$\uparrow$ & W@1$\uparrow$ \\
        \midrule
        Complete objective
        & \textbf{0.5585} & 0.3310 & \textbf{65.71\%} & \textbf{85.71\%} \\
        w/o Normal background
        & 0.5900 & 0.3647 & 57.14\% & \textbf{85.71\%} \\
        w/o all heatmap BCE
        & 0.8508 & 0.5336 & 48.57\% & 74.29\% \\
        w/o timestamp loss
        & 0.5607 & 0.3157 & 64.76\% & \textbf{85.71\%} \\
        w/o alignment
        & 0.6012 & 0.3808 & 60.00\% & 83.81\% \\
        \bottomrule
    \end{tabular}
    \vspace{-.1in}
\end{table}

\begin{table}[t]
    \caption{Auxiliary bidirectional retrieval for Full models under Crash-and-Normal training. Each direction contains 210 queries and 210 candidates. Values are Recall@K (\%).}
    \label{tab:retrieval_results}
    \vspace{-.1in}
    \centering
    \scriptsize
    \setlength{\tabcolsep}{3pt}
    \begin{tabular}{lrrrrrr}
        \toprule
        & \multicolumn{3}{c}{T2V} & \multicolumn{3}{c}{V2T} \\
        \cmidrule(lr){2-4}\cmidrule(lr){5-7}
        Backbone
        & R@1 & R@5 & R@10
        & R@1 & R@5 & R@10 \\
        \midrule
        EVA01
        & 35.71 & 72.86 & 88.10
        & 37.14 & \textbf{79.52} & \textbf{89.05} \\
        SigLIP2
        & \textbf{41.90} & \textbf{80.00} & \textbf{88.57}
        & \textbf{38.57} & 78.10 & 88.57 \\
        PE-Core
        & 35.24 & 76.67 & 87.62
        & 35.71 & 79.05 & 88.10 \\
        \bottomrule
    \end{tabular}
    \vspace{-.1in}
\end{table}

\noindent\textbf{Auxiliary Cross-modal Retrieval.} Table~\ref{tab:retrieval_results} presents auxiliary in-dataset retrieval results for Full models. SigLIP2 attains a T2V R@1 of 41.90\% and the highest observed R@1 in both directions. EVA01 attains the highest V2T R@5 and R@10, so the highest-scoring backbone depends on the retrieval direction in these runs. Class-conditioned captions, one caption per video, and the absence of an external baseline or free-query relevance labels preclude general-purpose, cross-dataset, or unrestricted semantic-retrieval claims. Because every row uses captions, the table does not isolate their incremental effect.

\section{Conclusion and Limitations}

FPV-CRASH provides a class-balanced dataset and fixed-protocol reference evaluation for offline Crash Moment Localization, supported by a heatmap-based model with video-only inference. In the fixed Crash-and-Normal setting, SigLIP2 Full obtains 0.5585s MAE, 0.3310s MedAE, and 65.71\% W@0.5 (a 50\% MAE reduction and a 28.6-point W@0.5 gain over the strongest zero-training prior); the three-seed comparisons do not show a reliable caption benefit across training compositions. Conclusions are limited by dataset scale and source/scene coverage, independent re-annotation covering only 53 Crash clips (10\% of the Crash set), unaudited class-conditioned captions, and a clip-disjoint but non-source-disjoint split; they do not establish unseen-source generalization. The ten-slot non-causal model is an offline reference, while sensitivity to sampling, heatmap settings, loss weights, and external domains remains only partially characterized. Future work will broaden coverage, validate captions and temporal annotations, and explore denser temporal representations. Predicted timestamps can anchor automatic extraction of short pre-/post-impact windows for offline analysis of crash footage. We deliberately do not frame this as automated incident reporting: the footage depicts property damage and third-party material, so evidentiary use carries privacy and legal obligations that this work does not address.

\textbf{Acknowledgement.} We used generative AI tools to create the illustration in Figure~\ref{fig:task_overview} (\textit{ChatGPT}, Image 2.5); the layout and content of the figure were specified by the authors, and all authors reviewed the generated image for correctness.


\bibliographystyle{IEEEtran}
\bibliography{references}

\end{document}